\documentclass{article}
\usepackage{spconf,amsmath,graphicx,indentfirst,url}
\usepackage{CJKutf8}

\newcommand{\Fig}[1]{Fig. \ref{fig:#1}} 
\newcommand{\Figure}[3]{\vspace{-0mm} \includegraphics[width=#1,clip]{#2.eps} \vspace{-2mm} \caption{#3} \vspace{-3mm} \label{fig:#2}}

\newcommand{\drawfig}[4]{ 
  \begin{figure}[#1]
  \begin{center}
  \Figure{#2}{#3}{#4}
  \end{center} 
  \end{figure}
}

\title{JSUT Corpus: Free Large-Scale Japanese Speech Corpus\\for End-to-End Speech Synthesis}

\name{Ryosuke Sonobe, Shinnosuke Takamichi, and Hiroshi Saruwatari}
\address{Graduate School of Information Science and Technology, The University of Tokyo, \\
3-7-1 Hongo Bunkyo-ku, Tokyo 133--8656, Japan}

\begin{document}
%
\maketitle
\begin{abstract}
	Thanks to improvements in machine learning techniques including deep learning, a free large-scale speech corpus that can be shared between academic institutions and commercial companies has an important role. However, such a corpus for Japanese speech synthesis does not exist. In this paper, we designed a novel Japanese speech corpus, named the ``JSUT corpus,'' that is aimed at achieving end-to-end speech synthesis. The corpus consists of 10 hours of reading-style speech data and its transcription and covers all of the main pronunciations of daily-use Japanese characters. In this paper, we describe how we designed and analyzed the corpus. The corpus is freely available online. 
\end{abstract}


\vspace{-1mm}
\section{Introduction}
	Thanks to developments in deep learning techniques, studies on speech have accelerated \cite{hinton12dnnasr,oord16wavenet,takamichi17moment,saito17asv}. In particular, in speech-to-text and text-to-speech research, end-to-end conversion from speech to text or from text to speech is an actively targeted task. Some studies on speech synthesis reported methods that do not use linguistic knowledge, e.g., no use of intermediate representations such as phonemes, in English, Spanish, and German \cite{wang17tacotron,sotelo17char2wav,watts12thesis}. However, it is known that natural language processing for Japanese is a more difficult task, e.g., semantic parsing and grapheme-to-phoneme conversion \cite{kubo14}. We expect that a Japanese speech corpus that is freely available would accelerate related research such as on end-to-end speech synthesis. However, there are no existing corpora, e.g., \cite{abe90}, for this purpose.
	
    In this paper, we describe the results of constructing a free, large-scale Japanese speech corpus, named the ``JSUT (Japanese speech corpus of Saruwatari Laboratory, the University of Tokyo) corpus.'' The corpus is designed to have all pronunciations of daily-use characters and individual readings in Japanese, which are not measured by conventional intermediate representation, such as phonemes and prosody. Also, it includes different-domain utterances, such as loanwords, and travel-domain and precedent utterances. We recorded 10 hours of speech data read by a native Japanese speaker and analyzed its linguistic and speech statistics. The corpus, including Japanese text and speech data, is freely available online \cite{jsut_corpus}.

\vspace{-1mm}
\section{Corpus design}	\vspace{-1mm}
	\subsection{Structures}
		To accelerate end-to-end research, the main purpose of the JSUT corpus is to cover all of the main pronunciations of daily-use Japanese characters, not to cover intermediate representations such as phonemes. The corpus includes the following nine sub-corpora. Their name is formatted as {\it [NAME][NUMBER]}. {\it [NUMBER]} indicates the number of utterances of the sub-corpus.
          \begin{itemize} \setlength{\itemsep}{-1pt}
              \item {\bf basic5000} ... utterances to cover all of the main pronunciations of daily-use Japanese characters. 
              \item {\bf countersuffix26}: utterances including individual readings of counter suffixes.
              \item {\bf loanword128}: utterances including loanwords, e.g., verbs or nouns. 
              \item {\bf utparaphrase512}: utterances for which a word or phrase of a piece of text is replaced with its paraphrase.
              \item {\bf voiceactress100}: para-speech for a free corpus of Japanese voice actresses \cite{voiceactresscorpus}.
              \item {\bf onomatopee300}: utterances including famous Japanese onomatopee (onomatopoeia).
              \item {\bf repeat500}: repeatedly spoken utterances.
              \item {\bf travel1000}: travel-domain utterances.
              \item {\bf precedent130}: precedent-domain utterances.
          \end{itemize}

	\subsection{Components}
    	We describe how we designed the nine sub-corpora below.
		\subsubsection{basic5000} 	
        	This is the main sub-corpus of the JSUT corpus. In Japanese, 2136 {\it kanji} characters ({\it kanji} are the logographic characters used in the modern Japanese writing system) are officially defined as daily-use characters \cite{joyokanji}, and each character has individual pronunciations consisting of its individual {\it kunyomi} (Chinese readings) and {\it onyomi} (Japanese readings). For example, we pronounce \begin{CJK}{UTF8}{min} ``一'' \end{CJK} ({\it one} in English) as ``ichi,'' ``itsu,'' ``hito,'' and ``hito (tsu).'' we collected 5000 sentences from Wikipedia \cite{wikipedia} and the TANAKA corpus \cite{tanaka_corpus} so that all pronunciations of the daily-use kanji characters could be covered. Some of the pronunciations cannot be found in these corpora, therefore, we manually made additional sentences to cover the remaining readings. 
    
    	\subsubsection{countersuffix26} \vspace{-1mm}
    		In Japanese, numerals cannot quantify nouns by themselves, and the pronunciation of the numerals changes depending on the suffix. For example, \begin{CJK}{UTF8}{min} ``二'' \end{CJK} (``two'' in English) is pronounced ``ni'' with \begin{CJK}{UTF8}{min} ``個'' (ko) \end{CJK} as the suffix and ``futa'' with \begin{CJK}{UTF8}{min} ``つ'' (tsu) \end{CJK}. We crowdsourced 26 sentences including such counter suffixes.
            
    	\subsubsection{loanword128} \vspace{-1mm}
        	Japanese sentences spoken daily have many loanwords, e.g., verbs and nouns, for example, \begin{CJK}{UTF8}{min} ``ググる (guguru)'' \end{CJK} is a verb meaning to {\it Google}, and \begin{CJK}{UTF8}{min} ``ディズニー (dyizunii)'' \end{CJK} means {\it Disney}. The pronunciations and accents of loanwords are a curious task in spoken language processing \cite{kubozono06loanwordaccent}. We crowdsourced such words and sentences. Also, we collected sentences from Wikipedia that included pronunciations not included in the modern Japanese system, for example,  sentences that had a Japanese-accented foreign proper name. 
            
    	\subsubsection{utparaphrase512} \vspace{-1mm}
    		Paraphrasing, e.g., lexical simplification, is a technique that substitutes a word or phrase into another sentence \cite{moku12simplification,kajiwara15snowe4}.
            It can support the reading comprehension of a wide range of readers in speech communication.
            The SNOW E4 corpus \cite{kajiwara15snowe4,snowe4corpus} includes sentences and a list of its paraphrased words.
            We chose one paraphrased word per sentence, and constructed 256 sentences and paraphrased sentences.
            The total number of sentences was 512.
            
    	\subsubsection{voiceactress100} \vspace{-1mm}
    		The Voice Actress Corpus \cite{voiceactresscorpus} is a free speech corpus of professional Japanese voice actresses and includes not only neutral but also emotional voices.
            Collecting para-speech for this speech corpus is very helpful to build attractive and emotional speech synthesis systems.
            We used sentences from this corpus and manually modified the pause positions.
            
    	\subsubsection{Onomatopee300} \vspace{-1mm}
    		Onomatopee (onomatopoeia) has an important role in connecting speech and non-speech sounds in nature, and Japanese is rich in onomatopoeia words. We crowdsourced 300 sentences having individual onomatopoeia words.
    
    	\subsubsection{repeat500} \vspace{-1mm}
    		Human speech production is not deterministic, i.e., speech waveforms always differ 
            even if we try to reproduce the same linguistic and para-linguistic information.
            Takamichi et al. \cite{takamichi17moment} proposed moment matching network-based speech synthesis 
            that synthesizes speech with natural randomness within the same contexts.
            To quantify randomness, we recorded utterances spoken repeatedly by a single speaker. 
            The speaker made utterances 5 times for each of the 100 sentences of the Voice Actress Corpus. 
        
        \subsubsection{travel1000 and precedent138} \vspace{-1mm}
        	We further constructed sentences whose domain differed from the above corpora. 
            1000 travel-domain sentences were collected from English-Japanese Translation Alignment Data \cite{jpenaligncorpus}.
            Also, 138 copyright-free precedent sentences were collected from \cite{courts_in_japan}.
            The words and phrases of the precedent sentences were significantly different from the above corpora, 
            but some sentences are too difficult to read. Therefore, we manually removed and modified these sentences to make reading easier.

\vspace{-1mm}
\section{Results of data collection}	\vspace{-1mm}
	\subsection{Corpus specs}
    	We hired a female native Japanese speaker and recorded her voice in our anechoic room. She was not a professional speaker but had experience working with her voice. The recordings were made in February, March, September, and October of 2017 for a few hours each day. The speaker made the recordings herself with our recording system. The speech data was sampled at 48 kHz. We used Lancers \cite{lancers} to collect several kinds of Japanese sentences. The total duration was 10 hours including small amounts of the non-speech region. The 16 bit/sample RIFF WAV format was used. Sentences (transcriptions) were encoded in UTF-8. 
        
        The distributed corpora included UTF-8-encoded sentences, 48-kHz speech, and recording information. Because the recording period was comparably long and the objective scores among the recording days varied as shown below, the recording information shows what day the speech data was recorded. The power of the speech data was normalized, but basically we made no additional modifications. Commas were added between breath groups. The positions of the commas were manually annotated.
        
	\subsection{Analysis}
    	We analyzed the linguistic and speech information of the constructed corpus. Note that not all of the data was used for the analysis to shorten the computation time.
        First, we counted the number of moras (sub-syllables) and words within one utterance by using MeCab \cite{kudo04mecab} and NEologd \cite{sato17neologd_nlp,sato15neologd}. The utterance length is the important factor in speech synthesis using the sequence-to-sequence mechanisms \cite{wang16attention,miyoshi17seq2seqvc}. \Fig{eps/num_moras} and \Fig{eps/num_words} show histograms of the moras and words, respectively. 
        As we can see, the corpus included a variety of lengths, from short utterances (a few words and moras) to long utterances ($70$ words and $140$ moras).
    
    	Next, we analyzed the changes in speech statistics per recording day. Speech data recorded during long periods causes objective and subjective differences among recording days \cite{kawai04ximera}. The Mean of log F0 was calculated for each recording day. F0 was extracted by using the WORLD analysis-synthesis system \cite{morise16world}. 
      	\Fig{eps/f0_stats} shows the result. There was no special tendency in the first half of the recordings, but we can see that the log F0 increased for the days of the second half.
      
        \drawfig{t}{\linewidth}{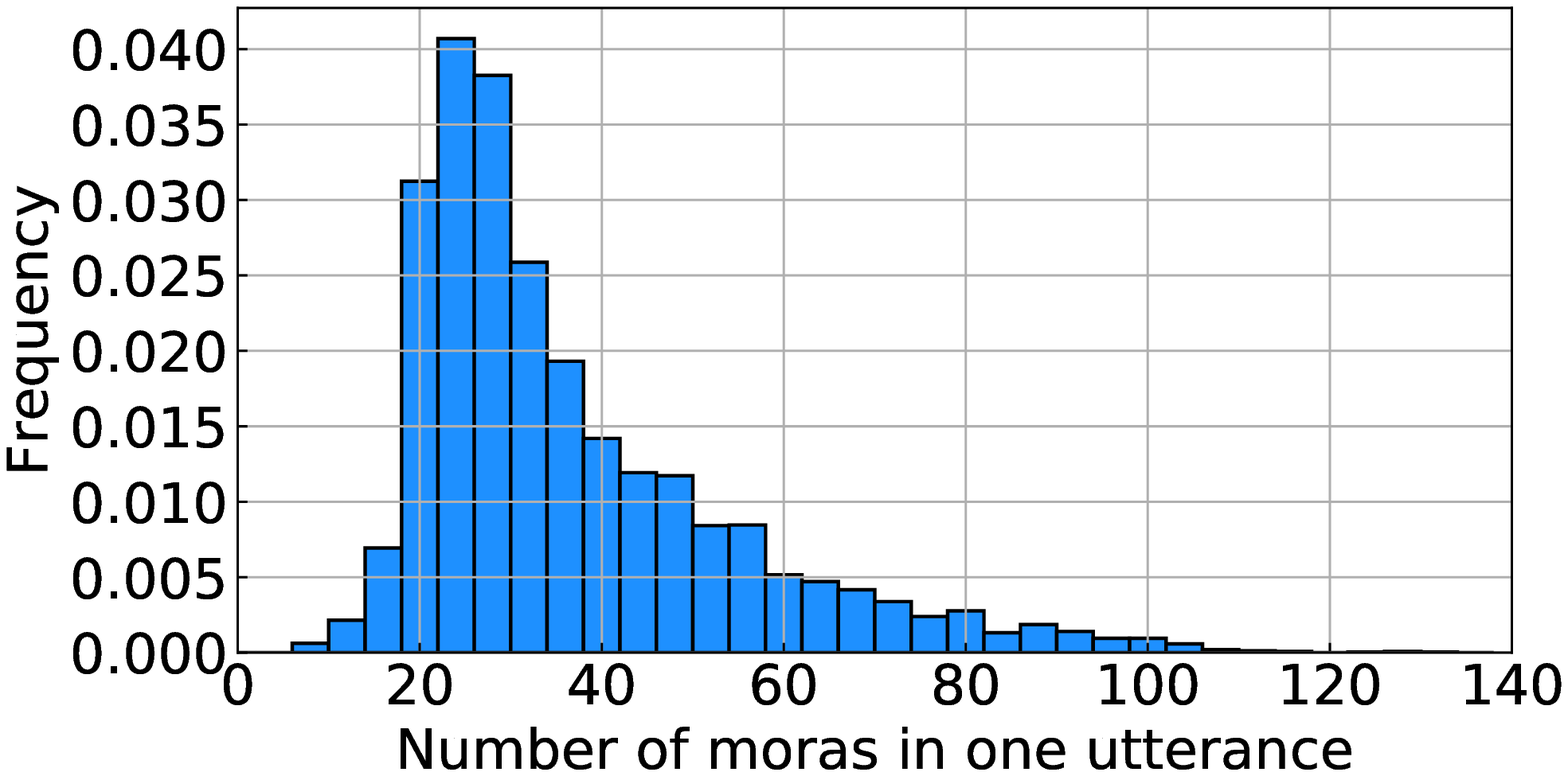}
        {Histogram of number of moras (sub-syllables) in one utterance. Minimum, mean, and maximum values are 7, 37.14, and 133, respectively. }
        
        \drawfig{t}{\linewidth}{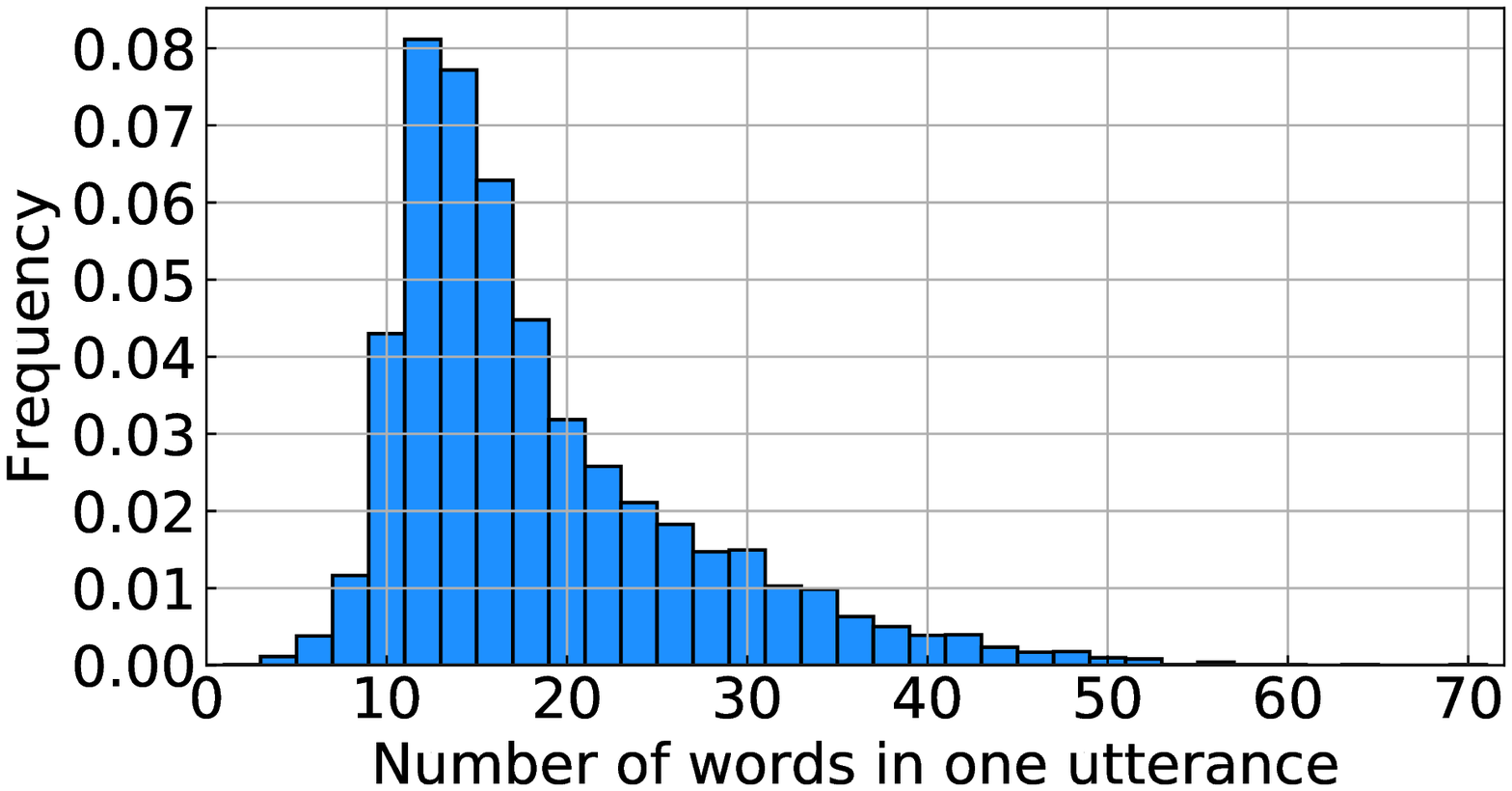}
        {Histogram of number of words in one utterance. Minimum, mean, and maximum values are 2, 18.03, and 70, respectively.}

        \drawfig{t}{\linewidth}{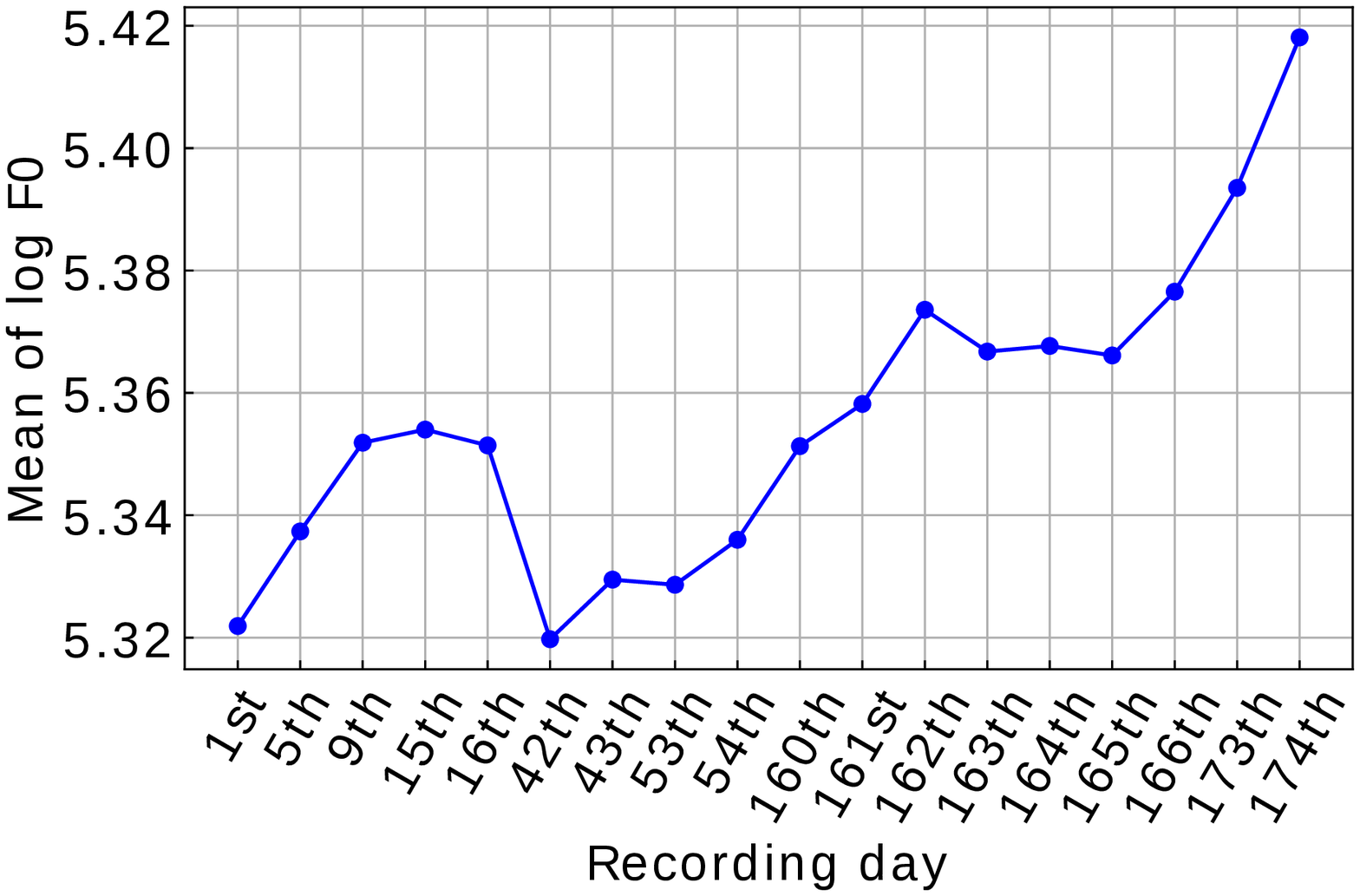}
        {Mean of log-scaled F0 for each recording day. Ordinal number of x-axis means how much time passed from ``1st'' recording day. For example, ``5th'' means 4 days after 1st recording day.}

\vspace{-1mm}
\section{Conclusion}	\vspace{-1mm}
	In this paper, we constructed a free, large-scale Japanese speech corpus (JSUT corpus) for end-to-end speech synthesis research. 
    The corpus was designed to have all pronunciations of daily-use kanji characters of Japanese and sentences of several domains.
    The corpus may be used for research by academic institutions and non-commercial research including research conducted within commercial organizations.

{\bf Acknowledgements:} Part of this work was supported by the SECOM Science and Technology Foundation. We thank Dr. Masahiro Mizukami of the Nara Institute of Science and Technology for the fruitful discussion on the paraphrase corpus,
Assistant Prof. Kazuhide Yamamoto of the Nagaoka University of Technology and
Tomoyuki Kajiwara of the Tokyo Metropolitan University for the use of the SNOW E4 corpus, and
the person in charge of the Voice Actress Corpus for the use of their corpus.

\bibliographystyle{IEEEbib}
\bibliography{tts}

\begin{thebibliography}{10}

\bibitem{hinton12dnnasr}
G.~Hinton, L.~Deng, D.~Yu, G.~Dahl, A.~r.~Mohamed, N.~Jaitly, A.~Senior,
  V.~Vanhoucke, P.~Nguyen, T.~Sainath, and B.~Kingsbury,
\newblock ``Deep neural networks for acoustic modeling in speech recognition:
  The shared views of four research groups,''
\newblock {\em Signal Processing Magazine of IEEE}, vol. 29, no. 6, pp. 82--97,
  2012.

\bibitem{oord16wavenet}
A.~v.~d. Oord, S.~Dieleman, H.~Zen, K.~Simonyan, O.~Vinyals, A.~Graves,
  N.~Kalchbrenner, A.~W. Senior, and K.~Kavukcuoglu,
\newblock ``Wave{N}et: {A} generative model for raw audio,''
\newblock vol. abs/1609.03499, 2016.

\bibitem{takamichi17moment}
S.~Takamichi, K.~Tomoki, and H.~Saruwatari,
\newblock ``Sampling-based speech parameter generation using moment-matching
  network,''
\newblock in {\em Proc. INTERSPEECH}, Stockholm, Sweden, Aug. 2017.

\bibitem{saito17asv}
Y.~Saito, S.~Takamichi, and H.~Saruwatari,
\newblock ``Training algorithm to deceive anti-spoofing verification for
  {DNN}-based speech synthesis,''
\newblock in {\em Proc. ICASSP}, Orleans, U.S.A., Mar. 2017.

\bibitem{wang17tacotron}
Y.~Wang, RJ~Skerry-Ryan, D.~Stanton, Y.~Wu, Ron~J. Weiss, N.~Jaitly, Z.~Yang,
  Y.~Xiao, Z.~Chen, S.~Bengio, Q.~Le, Y.~Agiomyrgiannakis, R.~Clark, and R.~A.
  Saurous,
\newblock ``Tacotron: Towards end-to-end speech synthesis,''
\newblock vol. abs/1609.03499, 2017.

\bibitem{sotelo17char2wav}
S.~Jose, M.~Soroush, K.~Kundan, S.~João F., K.~Kyle, C.~Aaron, and B.~Yoshua,
\newblock ``{{Char2Wav}: {E}nd-to-end speech synthesis},''
\newblock in {\em International Conference on Learning Representations
  (Workshop Track)}, April 2017.

\bibitem{watts12thesis}
O.~Watts,
\newblock ``Unsupervised learning for text-to-speech synthesis,''
\newblock {\em Ph. D thesis of the University of Edinburgh}, 2012.

\bibitem{kubo14}
K.~Kubo, S.~Sakti, G.~Neubig, T.~Toda, and S.~Nakamura,
\newblock ``Narrow adaptive regularization of weights for grapheme-to-phoneme
  conversion,''
\newblock in {\em Proc. ICASSP}, Florence, Italy, May 2014.

\bibitem{abe90}
M.~Abe, Y.~Sagisaka, T.~Umeda, and H.~Kuwabara,
\newblock ``{ATR} technical report,''
\newblock , no. TR-I-0166M, 1990.

\bibitem{jsut_corpus}
``{JSUT}: {J}apanese speech corpus of {S}aruwatari {L}ab, the {U}niversity of
  {T}okyo corpus,''
  \url{https://sites.google.com/site/shinnosuketakamichi/publication/jsut}.

\bibitem{voiceactresscorpus}
y\_benjo and MagnesiumRibbon,
\newblock ``Voice-actress corpus,'' \url{http://voice-statistics.github.io/}.

\bibitem{joyokanji}
Governments of~Japan Agency~for Cultural~Affairs,
\newblock ``List of daily-use kanjis
  \url{http://www.bunka.go.jp/kokugo_nihongo/sisaku/joho/joho/kijun/naikaku/kanji/index.html},''
\newblock 2010.

\bibitem{wikipedia}
``Wikipedia,'' \url{https://ja.wikipedia.org/}.

\bibitem{tanaka_corpus}
Y.~Tanaka,
\newblock ``Compilation of a multilingual parallel corpus,''
\newblock in {\em Proc. Pacling2001}, 2001.

\bibitem{kubozono06loanwordaccent}
H.~Kubozono,
\newblock ``Where does loanword prosody come from?: A case study of {J}apanese
  loanword accent,''
\newblock {\em Lingua}, vol. 116, no. 7, pp. 1140--1170, 2006.

\bibitem{moku12simplification}
M.~Moku, K.~Yamamoto, and A.~Makabi,
\newblock ``Automatic easy {J}apanese translation for information accessibility
  of foreigners,''
\newblock in {\em the Workshop on Speech and Language Processing Tools in
  Education}, 2012, pp. 85--90.

\bibitem{kajiwara15snowe4}
K.~Tomoyuki and Y.~Kazuhide,
\newblock ``Evaluation dataset and system for japanese lexical
  simplification,''
\newblock in {\em Proceedings of the ACL-IJCNLP 2015 Student Research
  Workshop}, Beijing, China, July 2015, pp. 35--40.

\bibitem{snowe4corpus}
``{SNOW E4}: evaluation data set of japanese lexical simplification,''
  \url{http://www.jnlp.org/SNOW/E4}, 2010.

\bibitem{jpenaligncorpus}
M.~Utiyama and M.~Takahashi,
\newblock ``English-japanese translation alignment data,''
  \url{http://www2.nict.go.jp/astrec-att/member/mutiyama/align/index.html},
  2003.

\bibitem{courts_in_japan}
``{COURTS IN JAPAN},'' \url{http://www.courts.go.jp/app/hanrei_jp/search1}.

\bibitem{lancers}
``Lancers \url{http://www.lancers.jp},''
\newblock .

\bibitem{kudo04mecab}
T.~Kudo, K.~Yamamoto, and Y.~Matsumoto,
\newblock ``Applying conditional random fields to {J}apanese morphological
  analysis,''
\newblock in {\em Proc. EMNLP}, Barcelona, Spain, Jul. 2004, pp. 230--237.

\bibitem{sato17neologd_nlp}
T.~Sato, T.~Hashimoro, and M.~Okumura,
\newblock ``Implementation of a word segmentation dictionary called
  mecab-ipadic-neologd and study on how to use it effectively for information
  retrieval (in {J}apanese),''
\newblock in {\em Proceedings of the Twenty-three Annual Meeting of the
  Association for Natural Language Processing}, 2017, pp. NLP2017--B6--1.

\bibitem{sato15neologd}
T.~Sato,
\newblock ``Neologism dictionary based on the language resources on the web for
  {M}ecab,'' 2015.

\bibitem{wang16attention}
W.~Wang, S.~Xu, and B.~Xu,
\newblock ``First step towards end-toend parametric {TTS} synthesis: Generating
  spectral parameters with neural attention,''
\newblock in {\em Proc. INTERSPEECH}, San Francisco, U.S.A., Sep. 2016, pp.
  2243--2247.

\bibitem{miyoshi17seq2seqvc}
H.~Miyoshi, Y.~Saito, S.~Takamichi, and H.~Saruwatari,
\newblock ``Voice conversion using sequence-to-sequence learning of context
  posterior probabilities,''
\newblock in {\em Proc. INTERSPEECH}, Stockholm, Sweden, Aug. 2017, pp.
  1268--1272.

\bibitem{kawai04ximera}
H.~Kawai, T.~Toda, J.~Ni, M.~Tsuzaki, and K.~Tokuda.,
\newblock ``{XIMERA}: a new {TTS} from {ATR} based on corpus-based
  technologies,''
\newblock in {\em Proc. SSW5}, Pittsburgh, USA, June 2004, pp. 179--184.

\bibitem{morise16world}
M.~Morise, F.~Yokomori, and K.~Ozawa,
\newblock ``{WORLD}: a vocoder-based high-quality speech synthesis system for
  real-time applications,''
\newblock {\em IEICE transactions on information and systems}, vol. E99-D, no.
  7, pp. 1877--1884, 2016.

\end{thebibliography}

\end{document}